\documentclass[10pt,twocolumn,letterpaper]{article}

\usepackage{cvpr}
\usepackage{times}
\usepackage{epsfig}
\usepackage{graphicx}
\usepackage{amsmath}
\usepackage{amssymb}

\usepackage{authblk}
\usepackage{algorithm}
\usepackage{algorithmicx}
\usepackage{algpseudocode}

\usepackage[breaklinks=true,bookmarks=false]{hyperref}

\cvprfinalcopy 

\def\cvprPaperID{9} 

\begin{document}

\title{Utilizing the Instability in Weakly Supervised Object Detection}


\makeatletter
\def\thanks#1{\protected@xdef\@thanks{\@thanks\protect\footnotetext{#1}}}
\makeatother
\newcommand*\samethanks[1][\value{footnote}]{\footnotemark[#1]}
\author[1,2,*]{Yan Gao\thanks{$^{*}$Equally-contributed.}}
\author[1,2,*]{Boxiao Liu}
\author[1,2]{Nan Guo}
\author[1,2]{Xiaochun Ye}
\author[2]{Fang Wan}
\author[1,2]{\\Haihang You}
\author[1,2,$\mathcal{y}$]{Dongrui Fan\thanks{$^\mathcal{y}$Corresponding author.}}

\vspace{-3pt}

\affil[1]{State Key Laboratory of Computer Architecture, Institute of Computing Technology, \authorcr Chinese Academy of Sciences, Beijing, China}
\affil[2]{University of Chinese Academy of Sciences, Beijing, China \authorcr {\tt\small \{gaoyan,liuboxiao,guonan,yexiaochun,youhaihang,fandr\}@ict.ac.cn, wanfang13@mails.ucas.ac.cn}}

\maketitle

\begin{abstract}

Weakly supervised object detection (WSOD) focuses on training object detector with only image-level annotations, and is challenging due to the gap between the supervision and the objective. Most of existing approaches model WSOD as a multiple instance learning (MIL) problem. However, we observe that the result of MIL based detector is unstable, i.e., the most confident bounding boxes change significantly when using different initializations. We quantitatively demonstrate the instability by introducing a metric to measure it, and empirically analyze the reason of instability. Although the instability seems harmful for detection task, we argue that it can be utilized to improve the performance by fusing the results of differently initialized detectors. To implement this idea, we propose an end-to-end framework with multiple detection branches, and introduce a simple fusion strategy. We further propose an orthogonal initialization method to increase the difference between detection branches. By utilizing the instability, we achieve 52.6\% and 48.0\% mAP on the challenging PASCAL VOC 2007 and 2012 datasets, which are both the new state-of-the-arts.
\end{abstract}

\section{Introduction}

Weakly supervised object detection (WSOD) has attracted intensive attention recently~\cite{song2014learning,bilen2014weakly,MFMIL,WSDDN,OICR,W2F,TS2C,MLEM,xianggang}. Unlike fully supervised object detection, WSOD aims at training detectors with only image-level annotations, which cost much less human labor than bounding boxes annotations.

\begin{figure}[t]
  \centering
  \includegraphics[width=0.9\linewidth]{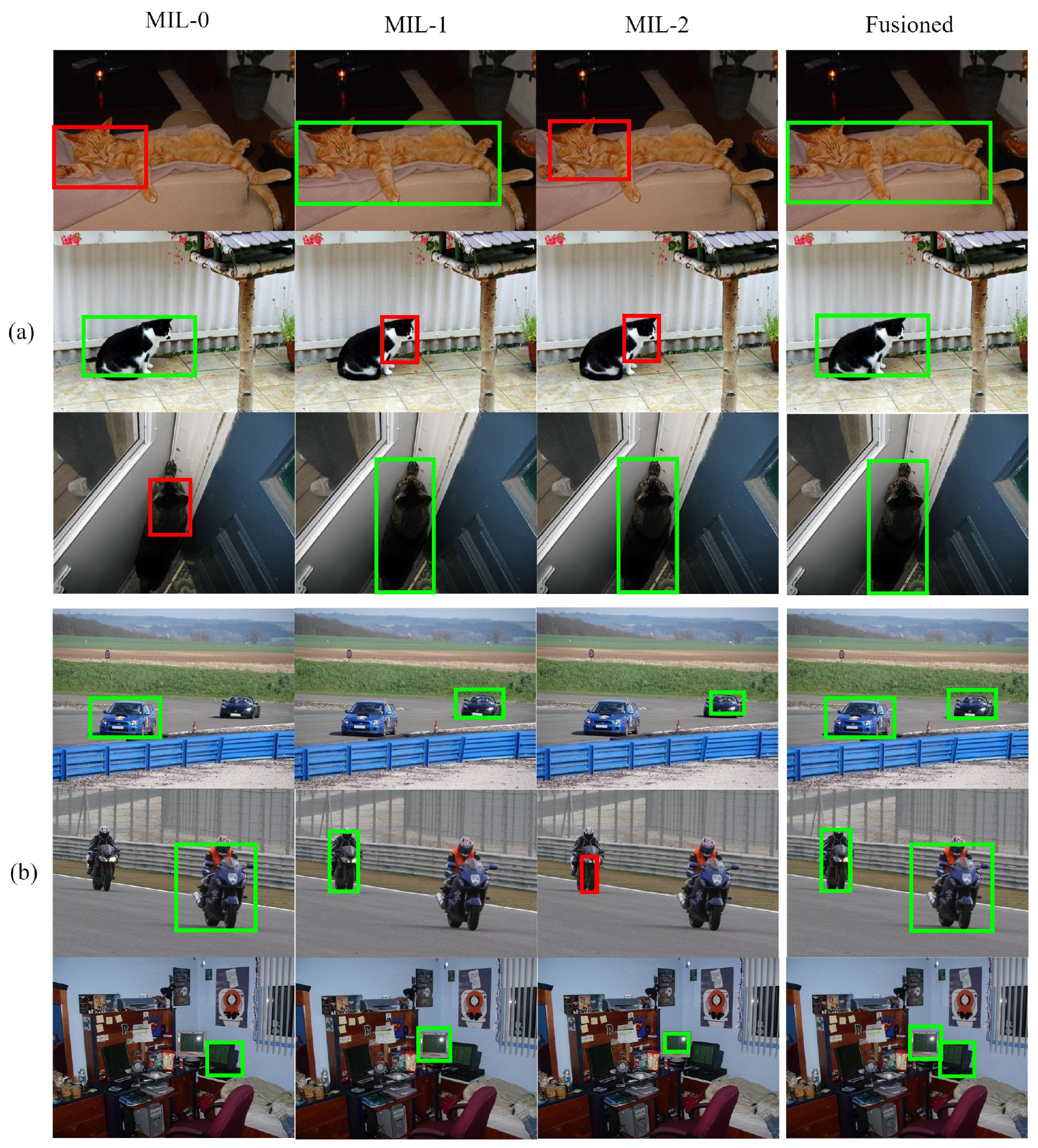}
  \caption{The instability of MIL-based detector. Each column corresponds to one MIL-based detector, and each row corresponds to one image. Green rectangles indicate the positive bounding boxes and red rectangles indicate the negative ones. (a) Images that contain only one object. (b) Images that contain multiple objects. Best viewed in color.}
  \label{fig:motivation}
  \vspace{-0.15in}
\end{figure}

A popular solution for WSOD is to formulate it as a multiple instance learning (MIL) problem. Training images are treated as labeled bags, which consist of multiple candidate bounding boxes. The learning procedure alternates between selecting the most confident proposals and using them to train a detector~\cite{MFMIL,song2014learning,SelfPaced}. Recently, many works combine convolutional neural networks (CNN) with MIL and get promising results~\cite{WSDDN,OICR,PCL,WSRPN,W2F,CL,TS2C,WCCN,MLEM,ZLDN,SelfTaught,xianggang}. Bilen \etal~\cite{WSDDN} propose a concise end-to-end Weakly Supervised Deep Detection Network (WSDDN), using two parallel branches to get classification and detection information. Based on WSDDN, some works propose to leverage regularization~\cite{WSDDN}, online refinement~\cite{OICR,PCL,WSRPN} and curriculum learning~\cite{ZLDN} to further improve the performance. Weakly supervised semantic segmentation is also introduced in WSOD to provide objectness information~\cite{WCCN,TS2C,xianggang}. However, there is still a remarkable performance gap compared with fully supervised detectors~\cite{Fast,Faster,Yolo,Ssd}.

We observe that the result of MIL-based detector is unstable. Specifically, detectors with different initializations may localize different regions on the same images. For example, the MIL-0 in the first column of Fig.~\ref{fig:motivation} (a) can correctly localize the cat in the image of the second row, but converge to the head of cat in the first and the third rows. However, MIL-1 in the second column of Fig.~\ref{fig:motivation} (a) succeeds to localize the cat in the first and third rows but fails in the second row. Also, if there are multiple objects in the image, detectors with different initial parameters may localize different one, Fig.~\ref{fig:motivation}(b).

The instability of MIL-based detector seems notorious as it limits the performance and leads to a high variance of the result, but we propose to utilize it on the contrary. Our motivation is that, by fusing the results of detectors with different initializations, we can keep the good candidate proposal and suppress the bad ones, so as to improve the detection performance. To implement this idea, we introduce a novel end-to-end framework to utilize the instability. Specifically, we design a multi-branch network structure consisting of a classification branch and multiple detection branches, inspired by WSDDN. The results of detection branches are coupled with the classification branch and the coupled results are further processed by a fusion strategy, called Surrounded Candidate Suppression (SCS). We further refine the fused result by training instance classifiers, following the popular practice ~\cite{OICR,PCL,WSRPN,TS2C,W2F}. Moreover, in order to further increase the difference between different detection results, we propose a novel category orthogonal parameter initialization method which makes the initialization parameters of the same category in different detection branches orthogonal.

To show the effectiveness of utilizing the instability, we conduct extensive experiments on the challenging PASCAL VOC 2007 and 2012 benchmarks. With the proposed framework, we obtain 52.6\% and 48.0\% mAP on VOC 2007 and VOC 2012 respectively, which are the new state-of-the-arts.

In summary, the contributions of this paper are listed as follows:

\begin{enumerate}
\item We analyze the instability of MIL based detector, providing quantitative evidence and empirical explanation. Based on the analysis, we propose to utilize the instability to improve detection performance by fusing the results of differently initialized detectors.
\item As training multiple detectors with different initialization is time consuming, we propose a simple but effective end-to-end framework and an online fusion strategy to utilize the instability. An orthogonal parameter initialization method is further proposed to increase the difference between detection branches.
\item The proposed framework significantly outperforms previous methods, and creates new state-of-the-arts both on PASCAL VOC2007 and VOC2012 datasets.
\end{enumerate}

\section{Related Work}

The majority of existing methods formulate WSOD as an MIL~\cite{MIL} problem. Under this formulation, a training image is seen as a bag of candidate proposals ~\cite{MFMIL}. Given the labels of bags, the objective of MIL is to train a classifier to correctly separate positive proposals from negative ones.

However, the loss function of MIL is non-convex, and the optimization of MIL is sensitive to initialization ~\cite{LTA,MFMIL,bilen2014weakly,song2014learning}. In order to solve this issue, some works introduce better initialization methods. Deselaers \etal~\cite{LTA} propose to initialize object locations based on the objectness score. Cinbis \etal~\cite{MFMIL} propose to split the training data into multi folds to escape local optima. Beyond generating better initialization, some works propose to smooth the optimization of MIL to alleviate the non-convexity problem. Bilen \etal~\cite{bilen2014weakly} introduce a smoothed version of MIL that softly labels object instances. Song \etal~\cite{song2014learning} propose to use Nesterov's smoothing technique in latent SVM model. The proposed method is also related to the non-convexity of MIL, but we propose to utilize the instability, which is partly caused by the non-convexity.

In recent years, many works combine deep convolutional neural networks with MIL and achieve promising results ~\cite{WSDDN,OICR,PCL,WSRPN,W2F,CL,TS2C,WCCN,MLEM,ZLDN,SelfTaught,xianggang}. The pretrained CNN models provide generic visual feature representations and reliable initialization for MIL. Bilen \etal~\cite{WSDDN} proposed a two-stream weakly supervised deep detection network (WSDDN), which can be trained with image-level labels in an end-to-end manner. Tang \etal~\cite{OICR} add into WSDDN several instance classifiers, and propose an online instance classifier refinement method. Wan \etal~\cite{MLEM} focus on reducing the randomness of localization during the optimization of network. Also, some works propose to leverage weakly supervised semantic segmentation to improve detection performance. Diba \etal~\cite{WCCN} generate candidate proposals based on the segmentation result and perform MIL among them. Wei \etal~\cite{TS2C} introduce two metrics to select reliable candidate proposals by using segmentation result.

Moreover, some researchers propose to combine fully supervised detectors with MIL based ones ~\cite{CL,W2F,GAN}. Wang \etal~\cite{CL} introduce a collaborative learning framework and a consistence-based loss to combine WSDDN and Faster-RCNN. Zhang \etal~\cite{W2F} propose to mine better pseudo ground truths from the result of MIL based detector to train a fully supervised detector. Shen \etal~\cite{GAN} use a generative adversarial learning framework to build the connection between fully and weakly supervised detectors.

The proposed framework consists of multiple detection branches and is related to ~\cite{cheng2018revisiting,cheng2018decoupled,li2019scale} that also propose to use multi-branch network to improve detection result. Cheng \etal~\cite{cheng2018revisiting} propose to add a decoupled classification refinement module to suppress hard false positive results. Li \etal~\cite{li2019scale} use multiple branches with different dilated convolution layers. Our method is different to these methods because the branches in our method have the same structure. Also, the branches in the proposed framework have different initializations in order to utilize the instability, which is not involved in these methods.

\begin{figure}[t!]
	\centering
	\includegraphics[width=0.9\linewidth]{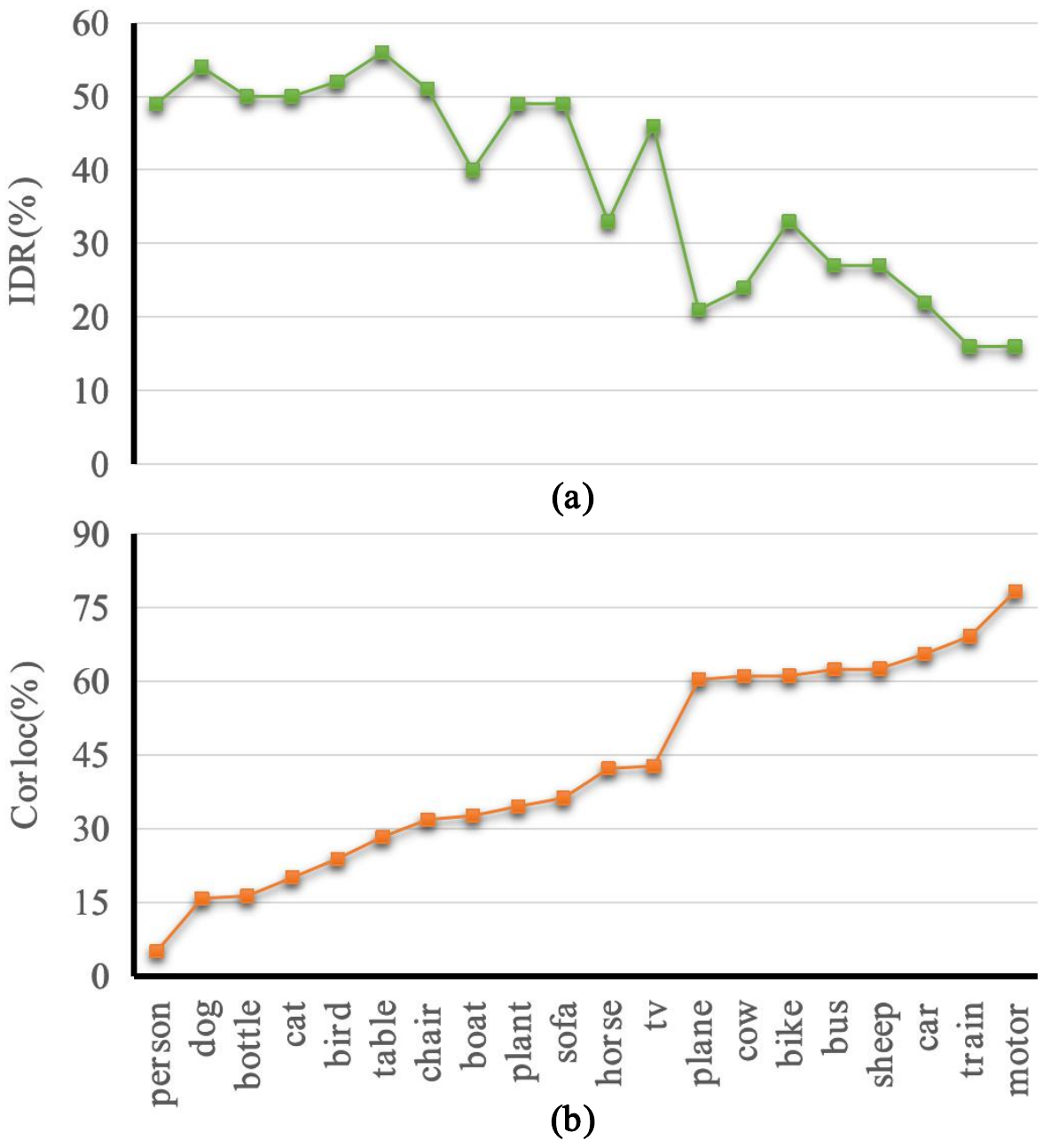}
	\caption{(a) Averaged CorLoc of 10 randomly initialized WSDDNs on each class. (b) Averaged IDR of each class, obtained by randomly sampling two WSDDNs 10 times.}
	\label{fig:idr}
	\vspace{-0.15in}
\end{figure}

\section{Analysis of Instability}
\label{section:analysis}

We observe that the result of MIL based detector shows significant instability, \emph{i.e.}, detectors trained with different initializations often localize different regions in the same image. To quantitatively analyze the instability of MIL based detectors, we first introduce a metric, Inconsistent Detection Rate (IDR), representing the inconsistency between the results of two detectors. In an image, if the IoU of the top scoring bounding boxes of two detectors is less than 0.5, we say the results are inconsistent. IDR indicates the rate of images where the results of two detectors are inconsistent. Formally, the IDR on class $c$ is defined as
\begin{equation}
IDR^c=\frac{|\{I^c_k, \text{ where }IoU(b^c_{1,k}, b^c_{2,k})< 0.5\}|}{|\{I^c_k\}|},
\label{eq:def_idr}
\end{equation}
where $I^c_k$ denotes the $k$th training image with positive label on class $c$, $b^c_{1,k}$ and $b^c_{2,k}$ denote the top scoring bounding boxes of two detectors on the $k$th image. The mean IDR over all classes is defined as
\begin{equation}
mIDR=\frac{\sum_{c=1}^{C}IDR^c}{C},
\label{eq:def_midr}
\end{equation}
where $C$ denotes the number of classes.

\begin{figure}[t]
  \centering
  \includegraphics[width=0.9\linewidth]{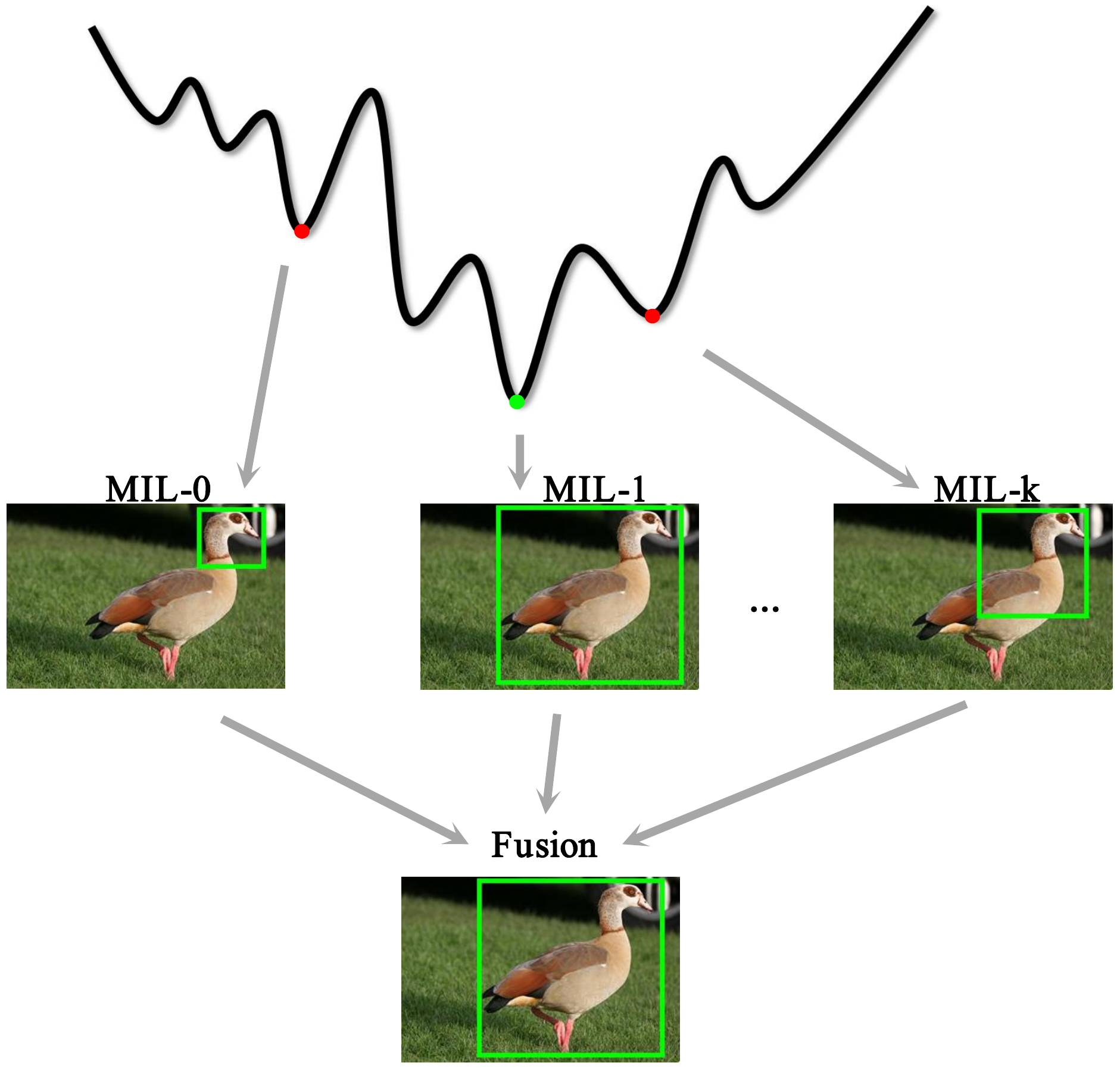}
  \caption{The non-convexity of MIL. Differently initialized detectors may be trapped in different local minimum, leading to different detection results. By fusing the results of different detectors, we may get more accurate localization of object.}
  \label{fig:non-convex}
  \vspace{-0.15in}
\end{figure}

We choose WSDDN, a popular MIL based detector, as a representative to show the instability problem. We train WSDDN for 10 times with different initial parameters. Then we randomly sample two detectors to compute IDR and mIDR for 10 times and show their averaged values. As shown in Fig.~\ref{fig:idr} (a), the mIDR reaches 38.3\% and the IDRs on some classes are even greater than 50\%, which means the detection result in about a half of images changes significantly if we change initial parameters of a detector. Also, the instability is more serious on classes with poor localization performance. From Fig.~\ref{fig:idr} (a) we can find that the class-specific IDR shows a negative relation with CorLoc.

\begin{figure*}[th]
	\centering
	\includegraphics[width=0.9\linewidth]{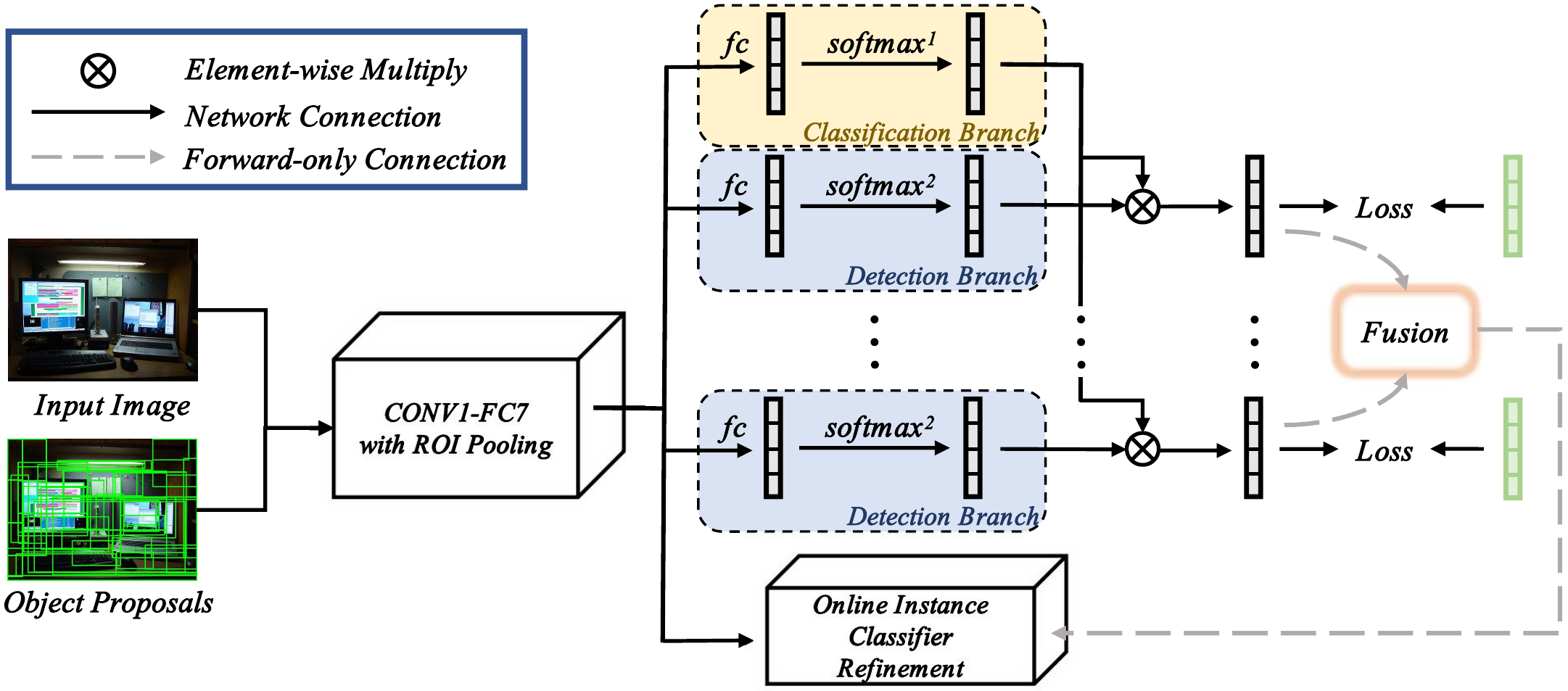}
	\caption{The network architecture of our method. A CNN backbone with ROI pooling and two fully connected layers is used to get the feature vectors of proposals. Then the feature vectors are fed into a classification branch, $K$ detection branches and several instance classifiers. $\text{Softmax}^1$ indicates softmax operation over the classes, and $\text{Softmax}^2$ indicates softmax operation over the proposals.}
	\label{fig:architecture}
	\vspace{-0.15in}
\end{figure*}

We think the reasons of this instability are two folds. Firstly, MIL is inherently non-convex and may have many local optimum. Secondly, the proposals in the same image have strong spatial relationship, which is related to the detection task. Specifically, negative proposals that only contain part of object always appear together with the positive proposals in the positive bags. However, such negative proposals never appear in a negative bag. Both of such negative proposals and positive proposals may correspond to local optimums of MIL, Fig.~\ref{fig:non-convex}. For example, an image containing a bird always has a positive proposal that contains the entire bird and some negative proposals that only contain the head of the bird, but images that contain such negative proposals are never labeled as negative. Without instance-level annotations, this spatial correlation between proposals usually makes the detector confused, leading to strong instability.

It seems that the instability is notorious as it limits the detection performance and results in unreliable localization. However, we argue that the instability can be utilized to improve detection performance on the contrary. Considering different top scoring proposals generated by randomly initialized detectors, some of them may be tight bounding boxes while others may be not. If we can fuse these proposals, keep the good ones and discard the bad ones, the detection performance will be improved, Fig.~\ref{fig:non-convex}.

\section{Method}

To utilize the instability of MIL based detector, we propose to fuse the results of differently initialized detectors. A natural way is to train the detector, such as WSDDN, for several times, and then fuse the results after all the training processes. However, this procedure is time-consuming. So we further propose a novel end-to-end network and an online fusion strategy to utilize the instability of MIL based detector. The overall network architecture is shown in Fig.~\ref{fig:architecture}. The main part of the network is a multi-branch structure consisting of a classification branch and multiple detection branches, which are initialized with different parameters. The results of all branches will be further fused and refined online. Moreover, in order to enlarge the difference between detection branches, we introduce an orthogonal initialization strategy for detection branches. We will elaborate the details of the proposed network, the online fusion strategy and the orthogonal initialization method in the rest of this section.

\subsection{Multi-branch Network Structure}

To utilize the instability of MIL, we need to generate different sets of localization information and fuse them to get better detection results. Inspired by WSDDN, we design a multi-branch network structure to implement this idea. Formally, given a training image $I$ and the corresponding candidate proposals $\mathcal{B}$ generated by selective search method ~\cite{SS}, we feed $I$ and $\mathcal{B}$ into a CNN backbone with ROI pooling, which is pretrained on ImageNet dataset. We use the output of FC7 as the features of candidate proposals. Then the network branch into a classification branch and $K$ detection branches. Each branch consists of a fully connected layer and a softmax layer. In the classification branch, the fully connected layer maps the features of proposals into a matrix $x^c\in\mathbb{R}^{C\times |\mathcal{B}|}$, where $C$ is the number of categories. Then a softmax operation along the class-axis is performed to get the classification scores of proposals
\begin{equation}
[\sigma_{cls}(x^c)]_{ij}=\frac{e^{x_{ij}^{c}}}{\sum_{n=1}^{C} e^{x_{nj}^{c}}}.
\label{eq:cls_score}
\end{equation}
In the $k$th detection branch, the matrix after fully connected layer $x^{d,k}\in\mathbb{R}^{C\times |\mathcal{B}|}$ is passed through a softmax operator along proposal-axis, defined as:
\begin{equation}
[\sigma_{det}(x^{d,k})]_{ij}=\frac{e^{x_{ij}^{d,k}}}{\sum_{n=1}^{|\mathcal{B}|} e^{x_{in}^{d,k}}}.
\label{eq:det_score}
\end{equation}
The detection scores of each detection branch will be further coupled with the classification scores of the same classification branch to get the final scores by an element-wise product, defined as:
\begin{equation}
\sigma^k=\sigma_{cls}(x^c)\odot \sigma_{det}(x^{d,k}).
\label{eq:coupled_score}
\end{equation}
For each detection branch, we can compute the image classification score $p_c$ on class $c$ with a summation over all proposals, defined as:
\begin{equation}
p^k_c=\sum_{n=1}^{|\mathcal{B}|}\sigma_{c,n}^k.
\label{eq:image_score}
\end{equation}
With the image label $\mathcal{Y}=\{y_1, y_2, \ldots, y_C\}$, the loss for $k$th detection branch $L_k$ and the total loss $L$ are defined as:
\begin{equation}
L^k=-\sum_{c=1}^C \{y_c\log p^k_c+(1-y_c)\log(1-p^k_c)\}
\label{eq:mil_loss}
\end{equation}
\begin{equation}
L=\sum_{k=1}^K L^k
\label{eq:total_loss}
\end{equation}

Instead of fusing the results of different detectors after the training procedure, we propose a simple online fusion strategy. After forward propagation in each training step, we can get the final scores of detection branches, and then fuse the results of different detection branches based on these scores. The fused results will be further refined by training instance classifiers. Here we follow ~\cite{OICR, PCL, WSRPN} to design the refinement module, which refines the fused results in a cascaded manner. For more details, please refer to ~\cite{OICR, PCL, WSRPN}.

\subsection{Online Fusion Strategy}
\label{section:fusion}

After getting the final scores of each detection branches, we introduce a simple online fusion strategy, called surrounded candidates suppression (SCS), to fuse the results of $K$ detection branches in every training step. The proposed strategy is based on the observation that WSDDN tends to localize the discriminative object parts or the whole object, rather than selecting bounding boxes that contain multiple objects or too many background regions. Thus, if a top scoring bounding box of a detector is surrounded by that of another detector, it is very likely that the surrounded bounding box only contains part of object and should be discarded. Formally, we first get the set of top scoring proposals of all detection branches $\mathcal{P}=\{P_1, P_2, \ldots, P_K\}$. Then we remove from $\mathcal{P}$ all proposals that are surrounded by another proposal in $\mathcal{P}$. Finally, a standard NMS with threshold 0.1 is performed among the remained proposals to get the fused result. With SCS, we can discard the bad candidate proposals and keep the good ones as many as possible. To make it clear, we summarize the process of SCS in Alg.~\ref{alg:SCS}.

\renewcommand{\algorithmicrequire}{\textbf{Input:}}  
\renewcommand{\algorithmicensure}{\textbf{Output:}}  

\begin{algorithm}[t!]
    \caption{Surrounded Candidates Suppression (SCS)}
    \begin{algorithmic}[1] 
    \Require The final score of $K$ detection branches $\{\sigma^1, \sigma^2,\ldots, \sigma^K\}$; object proposals $\mathcal{B}$; image label $\mathcal{Y}$.
    \Ensure Fused detection result $B_{fused}$.
    \State Set $B_{fused} \gets \{\}$.
    \For{$c=1$ \textbf{to} $C$}
        \If{$y_c=1$}
            \State Set $B_{top} \gets \{\}$.
            \For{$k=1$ \textbf{to} $K$}
                \State $b^k_c \gets \mathop{\arg\max}_{b_n\in B} \sigma_{c,n}^k$.
                \State $B_{top} \gets B_{top}\bigcup \{b^k_c\}$.
            \EndFor
            \For{$b^c_k \in B_{top}$}
                \If{$\exists b \in B_{top} \text{ s.t. } b^c_k \text{ is surrounded by } b$}
                    \State $B_{top} \gets B_{top}\setminus \{b^c_k\}$.
                \EndIf
            \EndFor
            \State $B^c_{fused} \gets \text{NMS}(B_{top}$).
            \State $B_{fused} \gets B_{fused}\bigcup B^c_{fused}$.
        \EndIf
    \EndFor
    \end{algorithmic}
    \label{alg:SCS}
\end{algorithm}

\begin{table*}[ht]
\begin{center}
\resizebox{1\textwidth}{!}{%
  \begin{tabular}{l | c c c c c c c c c c c c c c c c c c c c | c}
    \hline
    Method & aero & bike & bird & boat & bottle & bus & car & cat & chair & cow & table & dog & horse & mbike & person & plant & sheep & sofa & train & tv & mAP \\ \hline

    WSDDN~\cite{WSDDN} & 46.4&58.3&35.5&25.9&14.0&66.7&53.0&39.2&8.9&41.8&26.6&38.6&44.7&59.0&10.8&17.3&40.7&49.6&56.9&50.8&39.3\\

    OICR~\cite{OICR} & 58.5 & 63.0 & 35.1 & 16.9 & 17.4 & 63.2 & 60.8 & 34.4 & 8.2 & 49.7 & 41.0 & 31.3 & 51.9 & 64.8 & 13.6 & 23.1 & 41.6 & 48.4 & 58.9 & 58.7 & 42.0 \\

    WCCN~\cite{WCCN} & 49.5 & 60.6 & 38.6 & 29.2 & 16.2 & 70.8 & 56.9 & 42.5 & 10.9 & 44.1 & 29.9 & 42.2 & 47.9 & 64.1 & 13.8 & 23.5 & 45.9 & 54.1 & 60.8 & 54.5 & 42.8 \\
 
    TS2C~\cite{TS2C} &59.3 & 57.5 & 43.7 & 27.3 & 13.5 & 63.9 & 61.7 & 59.9 & 24.1 & 46.9 & 36.7 & 45.6 & 39.9 & 62.6 & 10.3 & 23.6 & 41.7 & 52.4 & 58.7 & 56.6 & 44.3 \\

    PCL~\cite{PCL} & 57.1 & 67.1 & 40.9 & 16.9 & 18.8 & 65.1 & 63.7 & 45.3 & 17.0 & 56.7 & 48.9 & 33.2 & 54.4 & 68:3 & 16:8 & 25.7 & 45.8 & 52.2 & 59.1 & 62.0 & 45.8 \\

    MLEM~\cite{MLEM} & 55.6 & 66.9 & 34.2 & 29.1 & 16.4 & 68.8 & 68.1 & 43.0 & 25.0 & 65.6 & 45.3 & 53.2 & 49.6 & 68.6 & 2.0 & 25.4 & 52.5 & 56.8 & 62.1 & 57.1 & 47.3 \\

    WSRPN~\cite{WSRPN} &60.3 & 66.2 & 45.0 & 19.6 & 26.6 & 68.1 & 68.4 & 49.4 & 8.0 & 56.9 & 55.0 & 33.6 & 62.5 & 68.2 & 20.6 & \textbf{29.0} & 49.0 & 54.1 & 58.8 & 58.4 & 47.9 \\ \hline

    OICR+FRCNN~\cite{OICR} & \textbf{65.5} & 67.2 & 47.2 & 21.6 & 22.1 & 68.0 & 68.5 & 35.9 & 5.7 & 63.1 & 49.5 & 30.3 & 64.7 & 66.1 & 13.0 & 25.6 & 50.0 & 57.1 & 60.2 & 59.0 & 47.0\\

    ZLDN~\cite{ZLDN} & 55.4 & 68.5 & \textbf{50.1} & 16.8 & 20.8 & 62.7 & 66.8 & 56.5 & 2.1 & 57.8 & 47.5 & 40.1 & 69.7 & 68.2 & 21.6 & 27.2 & 53.4 & 56.1 & 52.5 & 58.2 & 47.6 \\

    CL~\cite{CL} & 61.2 & 66.6 & 48.3 & 26.0 & 15.8 & 66.5 & 65.4 & 53.9 & 24.7 & 61.2 & 46.2 & 53.5 & 48.5 & 66.1 & 12.1 & 22.0 & 49.2 & 53.2 & 66.2 & 59.4 & 48.3 \\

    PCL+FRCNN~\cite{PCL} & 63.2 & 69.9 & 47.9 & 22.6 & 27.3 & \textbf{71.0} & 69.1 & 49.6 & 12.0 & 60.1 & 51.5 & 37.3 & 63.3 & 63.9 & 15.8 & 23.6 & 48.8 & 55.3 & 61.2 & 62.1 & 48.8 \\

    WSRPN+FRCNN~\cite{WSRPN} & 63.0 & 69.7 & 40.8 & 11.6 & \textbf{27.7} & 70.5 & \textbf{74.1} & 58.5 & 10.0 & \textbf{66.7} & \textbf{60.6} & 34.7 & \textbf{75.7} & 70.3 & \textbf{25.7} & 26.5 & 55.4 & 56.4 & 55.5 & 54.9 & 50.4 \\\hline

    Baseline(WSDDN+ODR) & 44.3 & \textbf{71.0} & 45.6 & 24.2 & 15.4 & 70.0 & 69.5 & 47.0 & 21.8 & 65.9 & 37.5 & 59.8 & 52.7 & 70.4 & 7.2 & 26.4 & \textbf{59.8} & 60.5 & 67.5 & 64.4 & 49.0 \\

    Ours & 63.4 & 70.5 & 45.1 & 28.3 & 18.4 & 69.8 & 65.8 & 69.6 & \textbf{27.2} & 62.6 & 44.0 & 59.6 & 56.2 & \textbf{71.4} & 11.9 & 26.2 & 56.6 & 59.6 & 69.2 & \textbf{65.4} & 52.0 \\ 

    Ours+FRCNN & 62.7 & 69.1 & 43.6 & \textbf{31.1} & 20.8 & 69.8 & 68.1 & \textbf{72.7} & 23.1 & 65.2 & 46.5 & \textbf{64.0} & 67.2 & 66.5 & 10.7 & 23.8 & 55.0 & \textbf{62.4} & \textbf{69.6} & 60.31 & \textbf{52.6} \\ \hline

\end{tabular}}
\end{center}
\caption{ Detection average precision (AP \%) on the PASCAL VOC 2007 test set. The upper part shows the results of weakly supervised detectors, and the second part shows the results of fully supervised detector trained by using the output of weakly supervised detectors as pseudo ground truth.}
\label{table:mAP_2007}
\end{table*}

\begin{table*}[ht]
\begin{center}
\resizebox{1\textwidth}{!}{%
  \begin{tabular}{l | c c c c c c c c c c c c c c c c c c c c | c}
    \hline
    Method & aero & bike & bird & boat & bottle & bus & car & cat & chair & cow & table & dog & horse & mbike & person & plant & sheep & sofa & train & tv & mAP \\ \hline

    WSDDN+context\cite{Context-WSDDN} & 64.0 & 54.9 & 36.4 & 8.1 & 12.6 & 53.1 & 40.5 & 28.4 & 6.6 & 35.3 & 34.4 & 49.1 & 42.6 & 62.4 & \textbf{19.8} & 15.2 & 27.0 & 33.1 & 33.0 & 50.0 & 35.3 \\
    
    WCCN~\cite{WCCN} & - & - & - & - & - & - & - & - & - & - & - & - & - & - & - & - & - & - & - & - & 37.9 \\
    
    OICR~\cite{OICR} & - & - & - & - & - & - & - & - & - & - & - & - & - & - & - & - & - & - & - & - & 37.9\\
    
    TS2C~\cite{TS2C} & 67.4 & 57.0 & 37.7 & 23.7 & 15.2 & 57.0 & 49.1 & 64.8 & 15.1 & 39.4 & 19.3 & 48.4 & 44.5 & 67.2 & 2.1 & 23.3 & 35.1 & 40.2 & 46.6 & 45.8 & 40.0 \\
    
    PCL~\cite{PCL} & 63.4 & 64.2 & 44.2 & 25.6 & 26.4 & 54.5 & 55.1 & 30.5 & 11.6 & 51.0 & 15.8 & 39.4 & 55.9 & 70.7 & 8.2 & 26.3 & 46.9 & 41.3 & 44.1 & 57.7 & 41.6 \\
    
    MLEM~\cite{MLEM} & - & - & - & - & - & - & - & - & - & - & - & - & - & - & - & - & - & - & - & - & 42.4 \\
    
    OICR+FRCNN~\cite{OICR} & - & - & - & - & - & - & - & - & - & - & - & - & - & - & - & - & - & - & - & - & 42.5 \\
    
    ZLDN~\cite{ZLDN} & 54.3 & 63.7 & 43.1 & 16.9 & 21.5 & 57.8 & \textbf{60.4} & 50.9 & 1.2 & 51.5 & \textbf{44.4} & 36.6 & 63.6 & 59.3 & 12.8 & 25.6 & 47.8 & \textbf{47.2} & 48.9 & 50.6 & 42.9 \\
    
    CL~\cite{CL} & 70.5 & 67.8 & 49.6 & 20.8 & 22.1 & 61.4 & 51.7 & 34.7 & 20.3 & 50.3 & 19.0 & 43.5 & 49.3 & 70.8 & 10.2 & 20.8 & 48.1 & 41.0 & \textbf{56.5} & 56.7 & 43.3 \\  
    
    WSRPN~\cite{WSRPN} & - & - & - & - & - & - & - & - & - & - & - & - & - & - & - & - & - & - & - & - & 43.4\\
    
    PCL+FRCNN~\cite{PCL} & 69.0 & \textbf{71.3} & \textbf{56.1} & 30.3 & 27.3 & 55.2 & 57.6 & 30.1 & 8.6 & 56.6 & 18.4 & 43.9 & \textbf{64.6} & 71.8 & 7.5 & 23.0 & 46.0 & 44.1 & 42.6 & 58.8 & 44.2 \\
    
    TS2C+FRCNN~\cite{TS2C} & \textbf{73.9} & 64.2 & 45.7 & \textbf{30.7} & 16.4 & \textbf{62.0} & 56.7 & \textbf{62.4} & 16.1 & 52.2 & 20.0 & 39.5 & 54.0 & 72.1 & 2.7 & 25.9 & 46.6 & 44.7 & 47.9 & 54.4 & 44.4 \\ 
    
    WSRPN+FRCNN~\cite{WSRPN} & - & - & - & - & - & - & - & - & - & - & - & - & - & - & - & - & - & - & - & - & 45.7 \\ \hline
    
    Ours & 72.7 & 68.8 & 51.6 & 29.4 & \textbf{29.1} & 60.3 & 58.0 & 59.0 & \textbf{22.6} & \textbf{61.9} & 22.4 & \textbf{52.3} & 59.8 & \textbf{74.0} & 7.2 & \textbf{28.1} & \textbf{53.4} & 33.5 & 54.5 & \textbf{60.7} & 48.0 \\ \hline
    
\end{tabular}}
\end{center}
\caption{Detection average precision (AP \%) on the PASCAL VOC 2012 test set.}
\vspace{-0.2in}
\label{table:mAP_2012}
\end{table*}

\subsection{Orthogonal Initialization}

The detection branches in the proposed network is supposed to be different with each other, so the fused results can be better than the results of original branches. The difference of detection branches comes from the randomness of initialization, which is not reliable. We argue that the proposed method can benefit from more significant difference between the initial parameters of detection branches. So we propose an orthogonal initialization method, making sure that the parameters of the fully connected layers of different detection branches are orthogonal with each other on every class.

Similar initialization methods have been proposed in ~\cite{ortho} to avoid the vanishment of gradient in recurrent neural networks. We follow the implementation in ~\cite{ortho} to design our orthogonal initialization method. For the $k$th detection branch, the parameters of the fully connected layer is denoted as a matrix $m^k\in \mathbb{R}^{l\times C}$, where $l$ denotes the length of the feature vector of a proposal. For each class $c$, we construct get an orthogonal matrix $q^c\in \mathbb{R}^{l\times K}$ by performing QR factorization on a random matrix. Then we assign the value of the $k$th column in $q$ to the $c$th column of $m^k$.

\section{Experiments}

\subsection{Datasets and Evaluation Metrics}

We evaluate our method on two widely used datasets, PASCAL VOC 2007 and 2012 ~\cite{PASCAL}. For each dataset, we use the \emph{trainval} set for training, and the \emph{test} set for testing. Only image-level labels are used to train the network.

For evaluation, we choose two kinds of measurements: 1) Average Precision (AP) and the mean of AP (mAP) on the \emph{test} set, following the standard PASCAL VOC protocol; 2) CorLoc ~\cite{corloc} on the \emph{trainval} set to evaluate the localization accuracy. Based on the PASCAL criterion, a bounding box is considered to be positive if it has an $IoU\ge 0.5$ with the ground-truth for both metrics.

\begin{figure}[t!]
    \centering
    \includegraphics[width=0.9\linewidth]{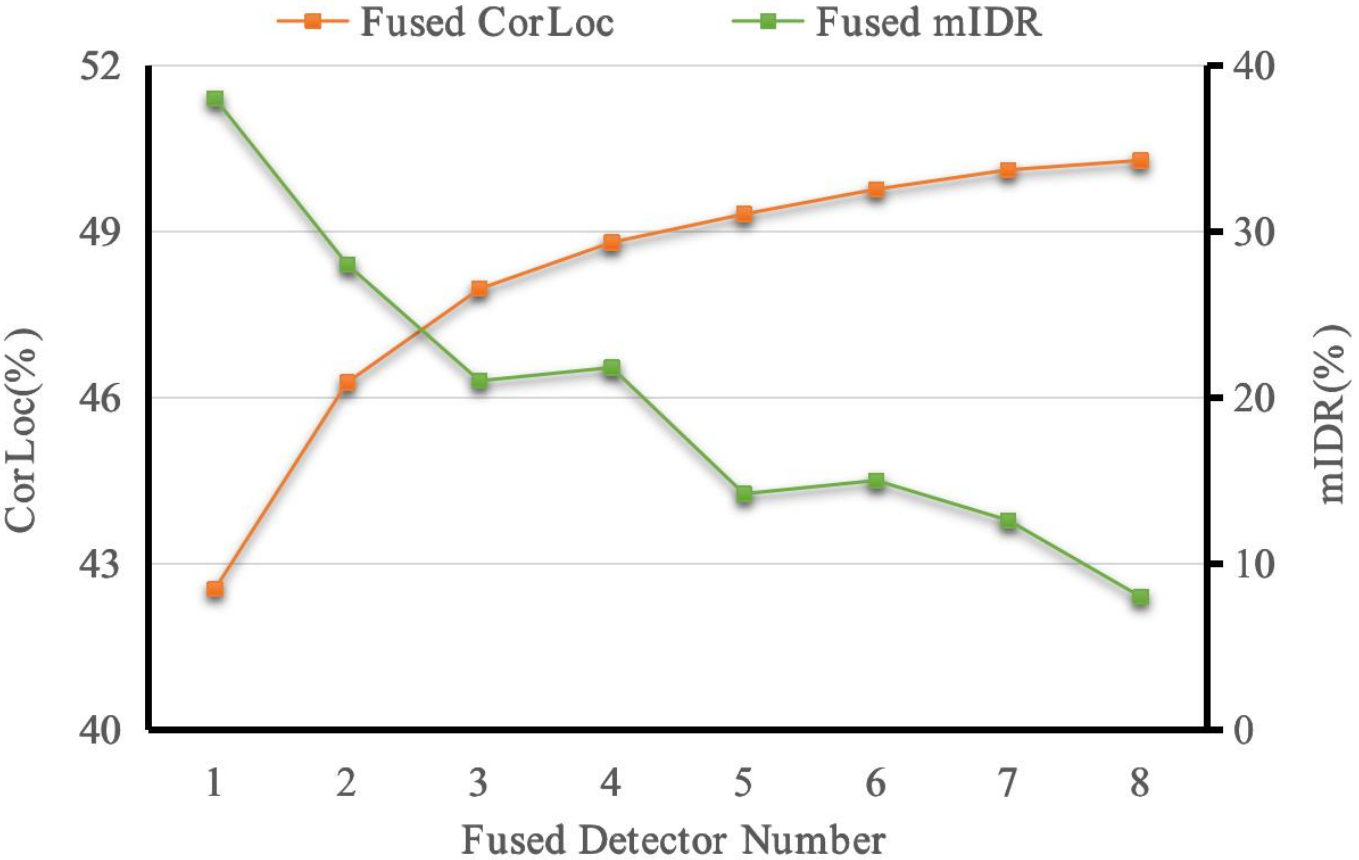}
    \caption{Averaged CorLoc and mIDR of fused results with different fused detector numbers $K$, obtained by randomly sampling $K$ WSDDNs 10 times.}
    \label{fig:fused_idr}
    \vspace{-0.15in}
\end{figure}

\subsection{Implementation Details}

We built our model on a VGG16~\cite{vgg16} network pretrained on ImageNet~\cite{ImageNet}. We remove the last fully connected layer, and replace the last max-pooling layer with an ROI pooling layer. The mini-batch for training is set to 2. The momentum and weight decay is set to 0.9 and $5\times 10^{-4}$ respectively. The learning rate is $5\times 10^{-4}$ for the first 10 epochs and then decrease to $5\times 10^{-5}$ for the following 5 epochs.

The image proposals are generated by selective search~\cite{SS}. For data augmentation, we use five image scales $\{480,576,688,864,1200\}$, with horizontal flips for both training and testing. In each training step, we randomly choose a scale to resize the image and then the image is randomly flipped. For all experiments, an NMS of 0.3 is employed to get final detection result. The average score of 10 augmented images is used as the final proposal scores. Our experiments are implemented based on PyTorch deep learning framework, and are conducted on NVIDIA GTX TitanX GPU.

\begin{table*}[ht]
\begin{center}
\resizebox{1\textwidth}{!}{%
  \begin{tabular}{l | c c c c c c c c c c c c c c c c c c c c | c}
    \hline
    Method & aero & bike & bird & boat & bottle & bus & car & cat & chair & cow & table & dog & horse & mbike & person & plant & sheep & sofa & train & tv & CorLoc \\ \hline

    WSDDN~\cite{WSDDN} &68.9 & 68.7 & 65.2 & 42.5 & 40.6 & 72.6 & 75.2 & 53.7 & 29.7 & 68.1 & 33.5 & 45.6 & 65.9 & 86.1 & 27.5 & 44.9 & 76.0 & 62.4 & 66.3 & 66.8 & 58.0\\

    OICR~\cite{OICR} & 85.4 & 78.0 & 61.6 & 40.4 & 38.2 & 82.2 & 84.2 & 46.5 & 15.2 & 80.1 & 45.2 & 41.9 & 73.8 & 89.6 & 18.9 & 56.0 & 74.2 & 62.1 & 73.0 & 77.4 & 61.2 \\

    WCCN~\cite{WCCN} &83.9 & 72.8 & 64.5 & 44.1 & 40.1 & 65.7 & 82.5 & 58.9 & 33.7 & 72.5 & 25.6 & 53.7 & 67.4 & 77.4 & 26.8 & 49.1 & 68.1 & 27.9 & 64.5 & 55.7 & 56.7 \\
 
    TS2C~\cite{TS2C} &84.2 & 74.1 & 61.3 & 52.1 & 32.1 & 76.7 & 82.9 & 66.6 & 42.3 & 70.6 & 39.5 & 57.0 & 61.2 & 88.4 & 9.3 & 54.6 & 72.2 & 60.0 & 65.0 & 70.3 & 61.0 \\

    PCL~\cite{PCL} & 81.7 & 82.4 & 63.4 & 41.0 & 42.4 & 79.7 & 84.2 & 54.9 & 23.4 & 78.8 & 54.4 & 46.0 & 75.9 & 89.6 & 22.8 & 51.3 & 72.2 & 66.1 & 74.9 & 76.0 & 63.0 \\

    MLEM~\cite{MLEM} & - & - & - & - & - & - & - & - & - & - & - & - & - & - & - & - & - & - & - & - & 61.4\\

    WSRPN~\cite{WSRPN} &81.2 & 81.2 & 60.7 & 36.7 & 52.3 & 80.7 & \textbf{89.0} & 65.1 & 20.5 & 86.3 & 61.6 & 49.5 & 86.4 & 92.4 & 41.4 & \textbf{62.6} & 79.4 & 62.4 & 73.0 & 75.6 & 66.9 \\ \hline

    OICR+FRCNN~\cite{OICR} & 85.8 & 82.7 & 62.8 & 45.2 & 43.5 & \textbf{84.8} & 87.0 & 46.8 & 15.7 & 82.2 & 51.0 & 45.6 & 83.7 & 91.2 & 22.2 & 59.7 & 75.3 & 65.1 & 76.8 & 78.1 & 64.3\\

    ZLDN~\cite{ZLDN} & 74.0 & 77.8 & 65.2 & 37.0 & 46.7 & 75.8 & 83.7 & 58.8 & 17.5 & 73.1 & 49.0 & 51.3 & 76.7 & 87.4 & 30.6 & 47.8 & 75.0 & 62.5 & 64.8 & 68.8 & 61.2 \\

    CL~\cite{CL} & 85.8 & 80.4 & \textbf{73.0} & 42.6 & 36.6 & 79.7 & 82.8 & 66.0 & 34.1 & 78.1 & 36.9 & 68.6 & 72.4 & 91.6 & 22.2 & 51.3 & 79.4 & 63.7 & 74.5 & 74.6 & 64.7 \\

    PCL+FRCNN~\cite{PCL} & 83.8 & 85.1 & 65.5 & 43.1 & 50.8 & 83.2 & 85.3 & 59.3 & 28.5 & 82.2 & 57.4 & 50.7 & 85.0 & 92.0 & 27.9 & 54.2 & 72.2 & 65.9 & 77.6 & \textbf{82.1} & 66.6 \\

    WSRPN+FRCNN~\cite{WSRPN} & 83.8 & 82.7 & 60.7 & 35.1 & \textbf{53.8} & 82.7 & 88.6 & 67.4 & 22.0 & 86.3 & \textbf{68.8} & 50.9 & \textbf{90.8} & \textbf{93.6} & \textbf{44.0} & 61.2 & 82.5 & 65.9 & 71.1 & 76.7 & 68.4 \\ \hline

    Baseline & 64.2 & 83.5 & 63.1 & 45.2 & 38.5 & 82.2 & 86.7 & 57.6 & 35.5 & 83.6 & 41.8 & 69.5 & 69.0 & 90.4 & 20.1 & 56.8 & 83.5 & \textbf{66.9} & 78.3 & 79.6 & 64.8 \\

    Ours & 84.2 & 84.7 & 59.5 & 52.7 & 37.8 & 81.2 & 83.3 & 72.4 & 41.6 & 84.9 & 43.7 & 69.5 & 75.9 & 90.8 & 18.1 & 54.9 & 81.4 & 60.8 & 79.1 & 80.6 & 66.9 \\ 

    Ours+FRCNN & \textbf{86.7} & \textbf{85.9} & 63.4 & \textbf{55.3} & 42.0 & \textbf{84.8} & 85.2 & \textbf{78.2} & \textbf{47.2} & \textbf{88.4} & 49.0 & \textbf{73.3} & 84.0 & 92.8 & 20.5 & 56.8 & \textbf{84.5} & 62.9 & \textbf{82.1} & 78.1 & \textbf{70.0} \\ \hline

\end{tabular}}
\end{center}
\caption{CorLoc on the \emph{trainval} set of VOC 2007.}
\label{table:CorLoc_2007}
\end{table*}

\begin{table}[th]
\begin{center}
\resizebox{0.3\textwidth}{!}{%
  \begin{tabular}{l | c }
    \hline
    Method & VOC 2012 \\ \hline
    
    WSDDN+context\cite{Context-WSDDN} & 54.8 \\
    
    ZLDN~\cite{ZLDN} & 61.5 \\
    
    OICR~\cite{OICR} & 63.5 \\
    
    TS2C~\cite{TS2C} & 64.4 \\
    
    PCL~\cite{PCL} & 65.0 \\
    
    CL~\cite{CL} & 65.2 \\  
    
    OICR+FRCNN~\cite{OICR} & 65.6 \\
    
    WSRPN~\cite{WSRPN} & 67.2 \\
    
    PCL+FRCNN~\cite{PCL} & 68.0 \\
    
    WSRPN+FRCNN~\cite{WSRPN} & \textbf{69.3} \\\hline
    
    Ours & 67.4 \\ \hline
\end{tabular}}
\end{center}
\caption{CorLoc on the \emph{trainval} set of VOC 2012.}
\vspace{-0.15in}
\label{table:CorLoc_2012}
\end{table}

\vspace{-1pt}

\subsection{The Effectiveness of Utilizing the Instability}

To demonstrate the effectiveness of utilizing the instability, we choose WSDDN as the basic detector, and fuse the results of WSDDNs that are initialized with different parameters. We use the same fusion strategy introduced in Section~\ref{section:fusion}, while we only keep the proposal with largest final score after NMS for the convenience of calculating CorLoc. As shown in Fig.~\ref{fig:fused_idr}, the CorLoc of the fused result increases monotonically as the number of fused detectors increases, and the instability of fused result decreases. Even by fusing two WSDDNs, the CorLoc increases from 42.54\% to 46.27\%, showing the effectiveness of utilizing the instability.

\subsection{Ablation Studies}

We first compared the proposed method with the baseline model, which combines WSDDN and the refinement module. Then we discuss the influence of detection branch number and the class-specific orthogonal initialization strategy. Without loss generality, we only conduct the ablation experiments on VOC 2007.

\vspace{-4pt}

\paragraph{Comparison with Baseline}

To show the effectiveness of the proposed framework, we compare the result of our method with a baseline framework containing a original WSDDN. As shown in Table.~\ref{table:mAP_2007}, our method improves the mAP from 49.0\% to 52.0\%. The performance on almost all classes has been improved, such as aeroplane (mAP from 44.3\% to 63.4\%), cat (mAP from 47.0\% to 69.6\%) and chair (mAP from 21.8\% to 27.2\%). Our model can mine more complete object bounding boxes by the fusion of multiple detectors while WSDDN can only find object parts. Also, the fusion strategy has the capacity of generating multiple candidate proposals with high confidence for the refinement module, further improving the detection performance.

\vspace{-4pt}

\begin{figure}[t!]
    \centering
    \includegraphics[width=0.9\linewidth]{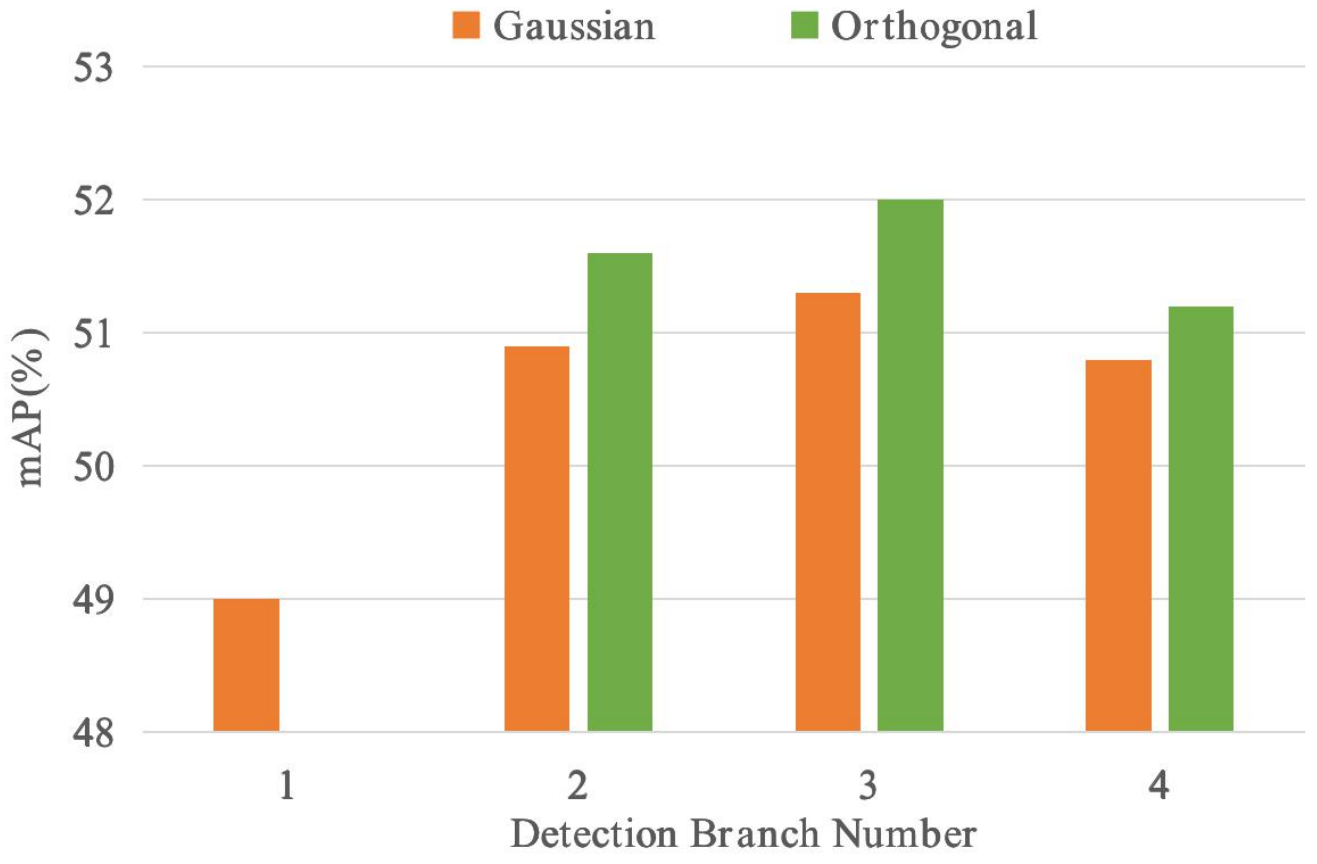}
    \caption{Results of different settings of branch number $K$ and initialization strategies. "Orthogonal" indicates the orthogonal initialization method. "Gaussian" indicates the Gaussian initialization method.}
    \label{fig:ablation_OI}
    \vspace{-0.15in}
\end{figure}

\paragraph{Influence of Detection Branch Number}
In Fig.~\ref{fig:ablation_OI}, we illustrate the result of ablation study on different numbers of detection branches. Even adding one more detection branch can significantly boost the performance (mAP from 49.0\% to 51.6\%), which confirms the effectiveness of our method. With 3 detection branches, the performance achieve the peak. The detection performance decreases slightly as the number of branches further increases. We think the reason may be that the online fusion strategy introduce the risk of localizing too big bounding boxes and this risk outweighs the gain of adding more branches when the number of detection branches is greater than 3. Thus, we set the detection branch number $K$ to 3 in other experiments.

\begin{figure*}[th]
    \centering
    \includegraphics[width=0.9\linewidth]{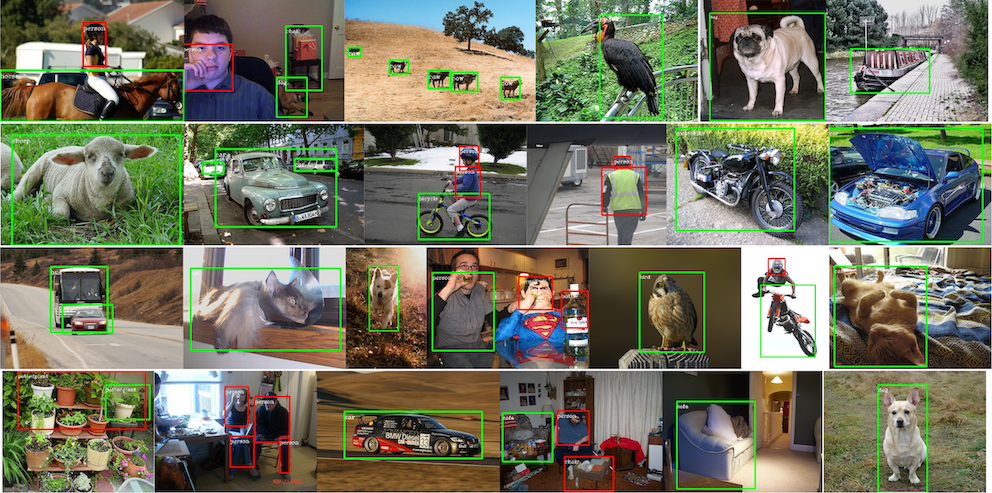}
    \caption{Some detection results of the proposed method. The green rectangles denote the correct detections (IoU $\ge$ 0.5), and the red rectangles denote the failed ones.}
    \label{fig:result}
    \vspace{-0.15in}
\end{figure*}

\paragraph{Influence of Orthogonal Initialization}

To validate the effectiveness of orthogonal initialization, we compare the proposed initialization method with a popular Gaussian initialization method, which samples values from Gaussian distribution to initialize the parameters. As shown in Fig.~\ref{fig:ablation_OI}, orthogonal initialization method significantly improves the detection performance. The effectiveness of orthogonal initialization further confirms the analysis in Section~\ref{section:analysis}.

\subsection{Comparison with State-of-the-Art}

In this subsection, we present the result of our framework compared with other state-of-the-art methods. Table.~\ref{table:mAP_2007} shows the result on VOC 2007 dataset, and Table.~\ref{table:mAP_2012} shows the result on VOC 2012 dataset. On VOC 2007, our model obtains 52.0\% mAP. On VOC 2012, our model obtains 48.0\% mAP. Our method outperforms previous methods with a large margin on both datasets.

Many works propose to train a fully supervised detector by using the result of MIL based detector as pseudo ground-truth, and show significant improvement of performance. Following Tang \etal~\cite{OICR}, we also use the top-scoring proposals produced by the proposed framework as pseudo ground-truth to train a Fast-RCNN~\cite{Fast}. As shown in Table.~\ref{table:mAP_2007}, the detection performance on VOC 2007 is further improved to 52.6\% in mAP, which is the new state-of-the-art. The CorLoc results of ours on VOC 2007 and VOC 2012 are reported in Table.~\ref{table:CorLoc_2007} and Table.~\ref{table:CorLoc_2012}, which also show the same trend.

We illustrate some detection results of our framework in Fig.~\ref{fig:result}. Although our model creates the new state-of-the-art, the detection results on some classes, such as person, chair and bottle, are still undesirable. The main failure for person is that the proposed method only finds different parts of person, such as face and hand. Although the inconsistency between the multiply detection branches is large, they all converge to object parts. As for indoor objects such as chairs and bottles, the co-occurrence of objects and backgrounds, or of different objects, is more common and makes it difficult to separate objects from contexts or from each other.

\section{Conclusions}

In this paper, we analyze the instability of MIL-based detector and introduce a metric IDR to measure the instability. Although the instability seems harmful, we propose to utilize it to get more accurate localization result. We propose an end-to-end network architecture and introduce an online fusion strategy to reduce computation cost. Also, a novel orthogonal initialization method is introduced to increase the difference between detection branches. Combined with refinement module, the proposed framework surpasses all previous methods and creates new state-of-the-art.

\section*{Acknowledgments}

This work was supported by the National Key Research and Development Program (2018YFB1003501, 2017YFB0202502), the National Natural Science Foundation of China (61732018, 61872335, 61802367), Austrian-Chinese Cooperative R\&D Project (FFG and CAS) Grant No. 171111KYSB20170032, the Strategic Priority Research Program of Chinese Academy of Sciences, Grant No. XDA18000000, and the Innovation Project Program of the State Key Laboratory of Computer Architecture (CARCH3303, CARCH3407, CARCH3502, CARCH3505).

{\small
\bibliographystyle{ieee_fullname}
\bibliography{egbib}
}

\end{document}